\documentclass{article}

\PassOptionsToPackage{numbers, compress}{natbib}

\usepackage[preprint]{neurips_2025}
\usepackage[utf8]{inputenc} 
\usepackage[T1]{fontenc}    
\usepackage{hyperref}       
\usepackage{url}            
\usepackage{booktabs}       
\usepackage{amsfonts}       
\usepackage{nicefrac}       
\usepackage{microtype}      
\usepackage{xcolor}         
\usepackage{graphicx}
\usepackage{amsmath}
\usepackage[ruled,vlined]{algorithm2e}
\usepackage{longtable}
\usepackage{wrapfig}
\usepackage{caption}
\usepackage{multirow}

\title{KG-BiLM: Knowledge Graph Embedding via Bidirectional Language Models}

\author{%
  Zirui Chen, Xin Wang\thanks{Corresponding author.}, Zhao Li, Wenbin Guo, Dongxiao He \\
  College of Intelligence and Computing\\
  Tianjin University\\
  \texttt{\{zrchen,wangx,lizh,wenff,hedongxiao\}@tju.edu.cn} \\
}

\begin{document}
\maketitle

\begin{abstract}
  Recent advances in knowledge representation learning (KRL) highlight the urgent necessity to unify symbolic knowledge graphs (KGs) with language models (LMs) for richer semantic understanding. However, existing approaches typically prioritize either graph structure or textual semantics, leaving a gap: a unified framework that simultaneously captures global KG connectivity, nuanced linguistic context, and discriminative reasoning semantics. To bridge this gap, we introduce \textbf{KG-BiLM}, a bidirectional LM framework that fuses structural cues from KGs with the semantic expressiveness of generative transformers. KG-BiLM incorporates three key components: (i) Bidirectional Knowledge Attention, which removes the causal mask to enable full interaction among all tokens and entities; (ii) Knowledge-Masked Prediction, which encourages the model to leverage both local semantic contexts and global graph connectivity; and (iii) Contrastive Graph Semantic Aggregation, which preserves KG structure via contrastive alignment of sampled sub-graph representations. Extensive experiments on standard benchmarks demonstrate that KG-BiLM\footnote{The source code of KG-BiLM is available at \url{https://anonymous.4open.science/status/kg-0317}} outperforms strong baselines in link prediction, especially on large-scale graphs with complex multi-hop relations—validating its effectiveness in unifying structural information and textual semantics.
\end{abstract}

\section{Introduction}
Recent advances in knowledge representation learning (KRL) highlight the need for unified approaches that integrate large-scale knowledge graphs (KGs) with structural representations and modern language models (LMs) with semantic understanding \cite{pan2023} \cite{mao2024}. In particular, significant progress in both symbolic and neural paradigms \cite{hwang2021} has motivated research beyond traditional knowledge graph embedding (KGE) methods to leverage contextual semantics more fully \cite{cambria2021} \cite{ji2020}. Moreover, the convergence of symbolic reasoning and neural text understanding has become increasingly prominent \cite{ye2021}. Consequently, investigating hybrid approaches that preserve KG structural relationships while capturing deep linguistic information is critical \cite{trajanoska2023} \cite{pan2023roadmap}.

\begin{figure}[htbp]
    \centering
    \includegraphics[width=0.93\textwidth]{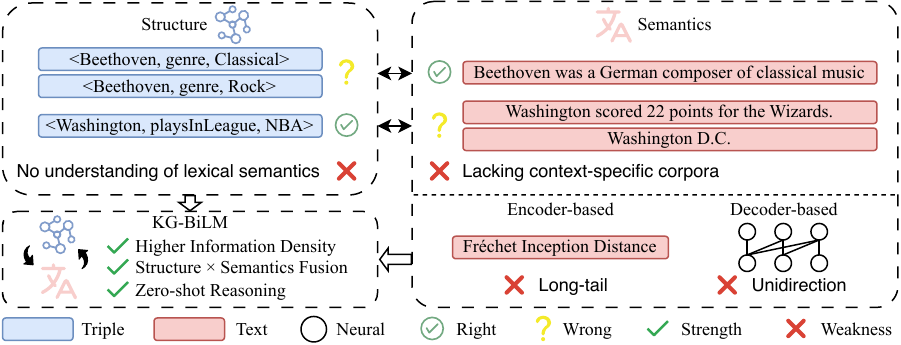}
    \caption{Illustration of complementary failure modes in KRL models. (a) Translation-based KGEs rate $\langle$Beethoven, genre, Classical$\rangle$ and $\langle$Beethoven, genre, Rock$\rangle$ almost identically—fixed offsets—while a BERT encoder uses context to prefer Classical. (b) In “Washington scored 22 points…,” LLaMA3-8B links Washington to the state; a graph encoder uses $\langle$Washington, playsInLeague, NBA$\rangle$ to pick the team. (c) Encoders falter on long-tail entities from scarce text, and decoder-only models can’t use future tokens to resolve earlier uncertainty.}
    \label{fig:motivation}
\end{figure}

Despite extensive research on KRL, the field remains fragmented. As shown in Fig.\ref{fig:motivation}, KGE methods, such as translation-based and semantic matching models, operate solely on symbolic triples and thus fail to capture lexical semantics, severely limiting their adaptability to heterogeneous downstream tasks \cite{yan2022}. Encoder-based transformers (e.g., BERT \cite{bert}) inject contextual word representations but are typically pre-trained on generic corpora, lacking the domain-specific text required for KG reasoning. Consequently, they struggle with long-tail entities and relations and cannot leverage the richer generative cues available to decoder-only models \cite{choi2021}. Conversely, applying decoder-based large language models (e.g., LLaMA \cite{llama2}) directly to KG tasks introduces new challenges. The inherent unidirectional generation paradigm obscures bidirectional dependencies and often ignores explicit structural constraints essential for faithful inference \cite{kopa2023}.

To bridge this research gap, we present a hybrid architecture called \emph{KG-BiLM}, aiming to (i) achieve higher information density than conventional KG encoders, (ii) integrate linguistic cues into KG representations, and (iii) enable zero-shot encoding capabilities. By overcoming the longstanding trade-offs between bidirectional structural reasoning, local semantic fidelity, and global topological consistency, KG-BiLM unifies the strengths of both structure- and semantics-oriented representation learning.

KG-BiLM unifies structural and semantic information via three core techniques: 
(1)~\textbf{Bidirectional Knowledge Attention} removes the causal mask from the decoder so that each token can attend to both past and future contexts, thereby strengthening inter-triple connections; 
(2)~\textbf{Knowledge-Masked Prediction} selectively masks triple and lexical tokens, forcing the model to leverage local lexical context alongside the global graph structure to reconstruct missing tokens, thereby integrating rich linguistic cues into the KG embeddings; and 
(3)~\textbf{Contrastive Graph Semantic Aggregation} applies contrastive learning on multiple graph-sampled views, aligning structural and semantic embeddings to preserve topology, enhance discriminative power, and facilitate zero-shot encoding of novel entities and relations. Together, these innovations transform a decoder-only transformer into a unified framework that jointly harnesses global KG structure and rich linguistic context for more accurate and generalizable reasoning.

The main contributions of this paper are as follows:
\begin{itemize}
    \item \textbf{Global Connectivity}: A novel causal masking scheme that enables bidirectional attention between tokens and entities along graph edges. This enhances multi-hop relational reasoning and captures long-range dependencies in a joint structural–textual context, thereby improving knowledge propagation with minimal overhead while preserving generative capabilities.
    \item \textbf{Contextual Inference}: An adaptive masking-recovery mechanism that hides and reconstructs entities and tokens using both graph adjacency and linguistic cues. This enforces structural–linguistic synergy, supports zero-shot generalization and robust missing-information recovery, and leverages a position-shifted loss to blend global and local signals effectively.
    \item \textbf{Semantic Discrimination}: A contrastive learning objective to align semantically similar graph–text pairs and repel dissimilar ones. This sharpens global embedding boundaries, ensures cluster cohesion and distinctive entity semantics at scale, and enhances retrieval precision under noisy, heterogeneous subgraph augmentations.
    \item \textbf{Comprehensive Evaluation}: We empirically demonstrate that KG-BiLM effectively preserves graph-structural cues while integrating textual semantics, establishing a new paradigm for holistic knowledge representation learning.
\end{itemize}

\section{Related Work} \label{sec:related}

\textbf{Knowledge Graph Embedding}. Translational (e.g., TransE \cite{transe}) and semantic-matching (e.g., DistMult \cite{distmult}) models dominate KGE, with numerous extensions that enrich multi-relational patterns and incorporate text, type, or logical signals for robustness \cite{NIPS2013_1cecc7a7,yang2015embeddingentitiesrelationslearning,Lin_Liu_Sun_Liu_Zhu_2015,10.5555/3045390.3045609,10.5555/3016100.3016172,toutanova-etal-2015-representing,10.5555/3171642.3171829,10.1007/978-3-319-63558-3_45}. Despite strong link-prediction accuracy, their fixed vectors rarely adapt across heterogeneous tasks, revealing the need for representations that unify structural fidelity with contextual flexibility.

\textbf{Encoder-based KRL}. Transformer encoders such as BERT \cite{bert}and RoBERTa \cite{roberta} are infused with KG-aware objectives to learn context-aligned entity embeddings \cite{devlin-etal-2019-bert,liu2019roberta,kg-bert,kepler,knn-kge}. While effective at capturing local text cues, their limited receptive fields overlook long-range graph signals, constraining zero-shot transfer and holistic structure-text integration.

\textbf{Decoder-based KRL}. Large language models (e.g., GPT-4o \cite{gpt4}, LLaMA \cite{llama2}) generate knowledge-grounded text or triples for downstream reasoning \cite{brown2020gpt3,touvron2023llama,kg-llm,cd,cp-kgc}. Their fluency supplies flexible supervision, yet the absence of explicit structural anchoring often weakens relational fidelity, underscoring the demand for methods that couple generative power with graph-consistent inference.

Due to the space limitation, please refer to Appendix \ref{appendix:related_work} for further details.

\section{The KG-BiLM Model} \label{sec:methodology}
This section introduces KG-BiLM, a knowledge graph–enhanced bidirectional large language model that unifies structural and textual semantics for robust knowledge representation.

\subsection{Preliminaries} \label{subsec:preliminary}
Let $\mathcal{G} = (\mathcal{E}, \mathcal{R}, \mathcal{T})$ be a knowledge graph, where $\mathcal{E}$ is the set of entities, $\mathcal{R}$ the set of relations, and $\mathcal{T}\subseteq\mathcal{E}\times\mathcal{R}\times\mathcal{E}$ the set of triplets $(e_i, r, e_j)$ describing relation $r\in\mathcal{R}$ between $e_i,e_j\in\mathcal{E}$. From pre-training corpora we derive a textual vocabulary $\mathcal{V}$. Our goal is to learn embeddings in a shared $d$-dimensional space so that every entity $e\in\mathcal{E}$ and every token $v\in\mathcal{V}$ is represented in $\mathbb{R}^d$, preserving both graph structure and linguistic context. Given a tokenized input sequence $\mathbf{x} = (x_1,\dots,x_N)$, which may intermix entity and relation names and natural-language tokens, the model parameters $\Theta$ map $\mathbf{x}$ to an initial embedding matrix $\mathbf{H}^{(0)}$. The full list of symbols is provided in Appendix \ref{appendix:natations}.


\subsection{Overview Framework} \label{subsec:framework}
Figure~\ref{fig:model} presents the overall architecture. KG-BiLM first encodes KG triples and entity description into a unified token sequence. These tokens are then processed by a Bidirectional Knowledge Attention module, which injects graph-aware masks. Next, we apply a Knowledge-Masked Prediction objective to reconstruct the masked tokens, leveraging in-batch negatives for more effective contrastive learning. Finally, a Contrastive Graph Semantic Aggregation step aligns paired views via an InfoNCE loss. The functions of the three modules (see Appendix \ref{appendix:modules} for full algorithmic details) are as follows:

\begin{figure}[htbp]
    \centering
    \includegraphics[width=\textwidth]{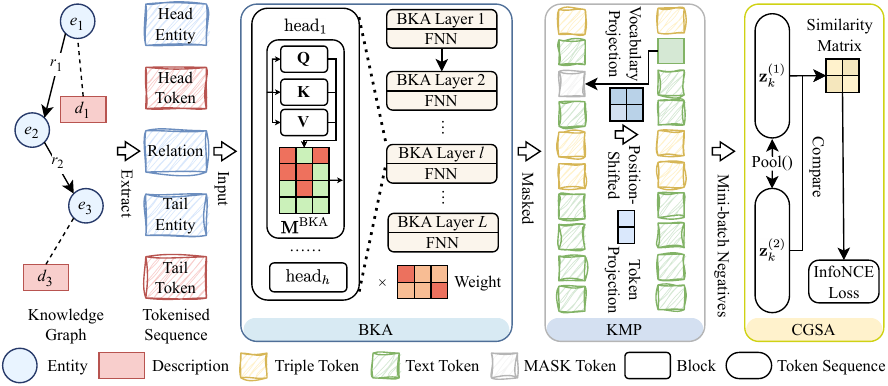}
    \caption{Overview of the KG-BiLM Architecture.}
    \label{fig:model}
\end{figure}

\begin{itemize}
\item \textbf{Bidirectional Knowledge Attention (BKA)} revises the causal mask of a decoder‑style transformer so that every position may attend to every other, strengthening entity–entity and token–token interactions without sacrificing autoregressive capabilities.
\item \textbf{Knowledge‑Masked Prediction (KMP)} strategically hides entities and textual tokens and trains the model to recover them using both the global graph topology and local linguistic context, thereby encouraging rich cross‑modal reasoning.
\item \textbf{Contrastive Graph Semantic Aggregation (CGSA)} samples semantically related sub‑graphs or textual variants and pulls their representations closer while pushing apart dissimilar samples, sharpening decision boundaries in the embedding space.
\end{itemize}


\subsection{Bidirectional Knowledge Attention} \label{subsec:attention}
Existing KGE methods typically rely on unidirectional or strictly local attention, which can hinder the learning of coherent global embeddings. Encoder-based approaches (e.g., BERT-like architectures) employ fully bidirectional attention but often lack the generative expressivity required for open‐ended inference. Decoder-based transformers, by contrast, use a causal mask that blocks access to future tokens, making it difficult to capture symmetric relations or long-range entity co‐occurrences. Consequently, a refined attention scheme is required to bridge these limitations. 

\textbf{Formulation of Bidirectional Knowledge Attention.} 
Formally, Let $L$ be the number of transformer layers, and let $\mathbf{H}^{(\ell)} \in \mathbb{R}^{N \times d}$ be the hidden states at the $\ell$-th layer. For layer $\ell$, define the query, key, and value matrices as:

\begin{equation}
\mathbf{Q} = \mathbf{H}^{(\ell)} \mathbf{W}_Q, \quad
\mathbf{K} = \mathbf{H}^{(\ell)} \mathbf{W}_K, \quad
\mathbf{V} = \mathbf{H}^{(\ell)} \mathbf{W}_V
\end{equation}

where $\mathbf{W}_Q, \mathbf{W}_K, \mathbf{W}_V \in \mathbb{R}^{d \times d_h}$ are trainable and $d_h$ denotes the dimension of each attention head (so that $h \times d_h = d$ for $h$ heads). In standard self-attention the mask $\mathbf{M}$  is set to $-\infty$ for positions $(i,j)$ if $j > i$ in a strictly causal decoder, or $0$ otherwise for a fully bidirectional encoder. Under the proposed BKA, a specialized mask $\mathbf{M}^{\mathrm{BKA}}$ is introduced:

\begin{equation}
\mathbf{M}^{\mathrm{BKA}}_{i,j} = 
\begin{cases}
0, & \text{if tokens }(x_i, x_j)\text{ may interact}\\
-\infty, & \text{otherwise}
\end{cases}
\label{eq:bkamask}
\end{equation}

The criterion ``may interact'' is determined not only by token positions but also by graph-based adjacency. Concretely, if $x_i$ and $x_j$ are entity tokens within the same sub-graph or share a relation path within a certain hop threshold, $\mathbf{M}^{\mathrm{BKA}}_{i,j}=0$. This ensures that each token may attend both backward and forward in the sequence, as well as to any closely connected entity in the KG.

Attention weights are then computed as:

\begin{equation}
\mathrm{Attention}(\mathbf{Q}, \mathbf{K}, \mathbf{V}; \mathbf{M}^{\mathrm{BKA}}) = 
\mathrm{softmax}\Big(\frac{\mathbf{Q}\mathbf{K}^\top}{\sqrt{d_h}} + \mathbf{M}^{\mathrm{BKA}} \Big) \mathbf{V}
\end{equation}

By permitting unrestricted forward and backward dependencies—while still enforcing structural constraints from the KG—the model captures richer, multi‐hop relational patterns. Consequently, BKA seamlessly fuses textual context with graph connectivity, alleviating the data sparsity often encountered in traditional KGE methods.

\textbf{Layer Update and Enhanced Representations.}
Denote the output of the $\ell$-th layer by $\mathbf{H}^{(\ell+1)}$. Employing multi-head attention, we obtain:

\begin{equation}
\tilde{\mathbf{H}}^{(\ell)} = \mathrm{Concat}\big(\mathrm{head}_1, \mathrm{head}_2,\dots, \mathrm{head}_h\big)\mathbf{W}_O
\end{equation}

where
\begin{equation}
\mathrm{head}_i = \mathrm{Attention}(\mathbf{Q}_i,\mathbf{K}_i,\mathbf{V}_i;\mathbf{M}^{\mathrm{BKA}})
\end{equation}

and $\mathbf{W}_O \in \mathbb{R}^{(h\cdot d_h)\times d}$. Then fuse the attended representations with a residual connection and layer normalization:

\begin{equation}
\bar{\mathbf{H}}^{(\ell)} = \mathrm{LayerNorm}\Big(\mathbf{H}^{(\ell)} + \tilde{\mathbf{H}}^{(\ell)}\Big)
\end{equation}

Next, a position-wise feed-forward network (FFN) is employed:

\begin{equation}
\mathbf{H}^{(\ell+1)} = \mathrm{LayerNorm}\Big(\bar{\mathbf{H}}^{(\ell)} + \mathrm{FFN}\big(\bar{\mathbf{H}}^{(\ell)}\big)\Big)
\end{equation}

This iterative transformation processes the sequence holistically, thereby allowing for the accumulation of signals from all tokens and their interlinked entities. The second advantage of using the multivariate approach is its capability to encode complex relational structures in a single pass, enhancing representational power for multiple downstream tasks.

By iterating this sequence of graph‐aware attention, residual fusion, and feed-forward updates, the model holistically aggregates signals from all tokens and their linked entities. This multivariate update not only captures complex relational structures in one pass but also substantially enhances the embedding power for downstream tasks.

\subsection{Knowledge-Masked Prediction} \label{subsec:masked}
Existing KRL pipelines often focus on direct link prediction or pairwise entity classification, which limits their ability to capture nuanced textual cues. In contrast, our method utilises knowledge-masked prediction (KMP) paradigm to gain a more holistic understanding of entity semantics. While KMP resembles traditional masked language modeling, it is specifically designed to incorporate relational paths and entity dependencies from the knowledge graph. This design makes the model robust to partial or missing observations—a key advantage for downstream zero-shot generalization.

\textbf{Notation for Masked Entities and Tokens.} 
Let $\mathbf{x}^m = (x^m_1, x^m_2,\dots, x^m_N)$ be the input sequence where a selected subset $\mathcal{M}\subset\{1,\dots,N\}$ of positions is masked. Each masked position $i\in \mathcal{M}$ could correspond to either an entity token or a textual token, replaced by a special mask symbol:

\begin{equation}
x^m_i = 
\begin{cases}
x_i, & \text{if } i \notin \mathcal{M}\\
\langle \mathrm{mask} \rangle, & \text{if } i \in \mathcal{M}
\end{cases}
\end{equation}

We denote the masking ratio by $\gamma = \frac{|\mathcal{M}|}{N}$, which can be tuned to regulate the difficulty of the inference task. 

\textbf{Predicting Masked Tokens and Entities.}
Once the BKA is applied, the hidden states $\mathbf{H}^{(L)}$ at the final layer encapsulate both local lexical context and global KG structure. Let $\mathbf{h}_i \in \mathbb{R}^d$ denote the final hidden representation at position $i$. The model then projects $\mathbf{h}_i$ into a distribution over the vocabulary $\mathcal{V}$ extended with entity symbols:

\begin{equation}
p_{\Theta}(x_i\mid \mathbf{h}_i) = \mathrm{softmax}\big(\mathbf{W}_P \mathbf{h}_i + \mathbf{b}_P\big)
\end{equation}

where $\mathbf{W}_P \in \mathbb{R}^{|\mathcal{V}+\mathcal{E}|\times d}$ and $\mathbf{b}_P \in \mathbb{R}^{|\mathcal{V}+\mathcal{E}|}$ are projection parameters. For each $i \in \mathcal{M}$, the training objective focuses on predicting the true token or entity $x_i$ based on the modified input $\mathbf{x}^m$. 

\textbf{Position-Shifted Loss Assignment.} 
A key innovation is that the model must leverage both preceding and succeeding context, as well as linked entities, to decode a masked position. In a strict causal decoder, the cross-entropy loss at position $i$ would typically be computed using the logits from $\mathbf{h}_i$. However, to enforce a deeper structural dependence, the present approach employs a position-shifted mechanism:

\begin{equation}
\mathcal{L}_{\mathrm{KMP}} = - \sum_{i \in \mathcal{M}} \log \, p_{\Theta}\Big(x_i \mid \mathbf{h}_{i-1}\Big)
\label{eq:kmp}
\end{equation}

where $\mathbf{h}_{i-1}$ denotes the hidden representation from the immediately preceding position. This position-shifted KMP scheme compels the network to aggregate contextual information from both preceding and succeeding tokens, as well as linked entities. It offers three key benefits: it enforces consistency across consecutive positions, encourages global attention to relational edges, and ensures that local neighborhoods in the sequence capture relevant semantically related tokens.

\textbf{Link to Zero-Shot Capability.} 
By repeatedly training the model to infer masked entity tokens with partial observations, it learns to exploit both cross-position and cross-relational dependencies. Consequently, the learned embeddings become adaptable to new entities and unseen textual patterns. The synergy between the BKA and the position-shifted KMP modules thus underpins robust zero-shot encoding: once structural and linguistic patterns are internalized, the model can generalize to novel contexts with minimal or no additional training data.

\subsection{Contrastive Graph Semantic Aggregation} \label{subsec:contrastive}
Despite the effectiveness of bidirectional attention and knowledge masking in unifying textual and graph-based cues, there remains a risk that entity embeddings become overly entangled with their local context and lose global distinctiveness. To mitigate this, we introduce a contrastive learning module—\emph{Contrastive Graph Semantic Aggregation} (CGSA)—that preserves discriminative power while still benefiting from integrative representations. The contrastive objective naturally clusters semantically or relationally coherent samples and pushes apart incongruent pairs, thereby reinforcing the structural integrity of the knowledge-graph representation.

\textbf{Sampling Mechanism and Dropout Variation.} 
To achieve contrastive alignment, we draw two independent corrupted views $\mathbf{x}^{(1)}$ and $\mathbf{x}^{(2)}$ from the same original sequence or sub-graph. Each view is generated by applying a separate random dropout mask. Specifically, for a given sub-graph or textual snippet $s$, two corrupted views $s_1$ and $s_2$ are generated through random dropout or data augmentation, yielding token sequences $\mathbf{x}^{(1)}$ and $\mathbf{x}^{(2)}$. Both are passed through the aforementioned BKA and KMP modules, resulting in final embeddings:

\begin{equation}
\mathbf{z}^{(1)} = \mathrm{Pool}\Big(\mathbf{H}^{(L)}(\mathbf{x}^{(1)})\Big), \quad
\mathbf{z}^{(2)} = \mathrm{Pool}\Big(\mathbf{H}^{(L)}(\mathbf{x}^{(2)})\Big)
\end{equation}

where $\mathrm{Pool}(\cdot)$ (e.g., mean pooling or a special classification token) aggregates the sequence‐level embedding.

\textbf{Contrastive Objective.}
Denote a minibatch by $\{\mathbf{z}^{(1)}_k, \mathbf{z}^{(2)}_k\}_{k=1}^B$, where each index $k$ corresponds to a distinct pair derived from the same original sub-graph $s_k$. Let $\mathrm{sim}(\mathbf{u}, \mathbf{v}) = \frac{\mathbf{u}^\top \mathbf{v}}{\|\mathbf{u}\|\|\mathbf{v}\|}$ measure the cosine similarity. Inspired by InfoNCE \cite{oord2018representation}, we formulate the CGSA loss as:

\begin{equation}
\mathcal{L}_{\mathrm{CGSA}} = - \sum_{k=1}^B \log \Bigg( \frac{\exp\big(\mathrm{sim}(\mathbf{z}^{(1)}_k,\, \mathbf{z}^{(2)}_k)\,/\,\tau\big)}{\sum_{\ell=1}^B \exp\big(\mathrm{sim}(\mathbf{z}^{(1)}_k,\, \mathbf{z}^{(2)}_\ell)\,/\,\tau\big)} \Bigg)
\label{eq:cgsa}
\end{equation}

where $\tau$ is a temperature hyperparameter. The numerator encourages high similarity between the two views of the same entity–text pair, while the denominator discourages alignment with representations of other pairs in the minibatch. By maximizing the similarity of paired views and minimizing the similarity with respect to non-matching samples, CGSA promotes cluster cohesion for semantically similar embeddings and cluster separation for dissimilar embeddings.

\textbf{Integration with Global KG Semantics.} 
The CGSA excels at preserving global semantics in large‐scale knowledge graphs with complex relational patterns. By enforcing that embeddings remain discriminative under repeated dropout corruptions, it also fosters stable representations across paraphrases or reordered textual segments describing the same entity. For example, if two sub-graphs $s_i$ and $s_j$ share the majority of their entities but differ in minor textual descriptions, the contrastive loss anchors their embeddings to be closer than those of unrelated sub-graphs. In doing so, CGSA addresses a key limitation of earlier generative LM-based methods, wherein structural coherence was often overshadowed by purely textual factors.

\section{Experiments} \label{sec:experiments}
This section empirically evaluates KG-BiLM on four widely used knowledge graph benchmarks, comparing it against state-of-the-art symbolic, neural and hybrid alternatives. We first detail the experimental setup, then analyse performance on structure-only and semantically-enriched datasets, followed by a component ablation, a zero-shot study, and a qualitative analysis.

\subsection{Experimental Settings} \label{subsec:exp_settings}

\begin{wraptable}{r}{0.53\textwidth}
  \centering
  \setlength{\tabcolsep}{1.5pt}
  \caption{Statistics of the KG benchmarks.}
  \scalebox{0.82}{
    \begin{tabular}{lccccccc}
    \toprule
    \textbf{Dataset} & \textbf{\#Ent.} & \textbf{\#Rel.} & \textbf{Train} & \textbf{Valid} & \textbf{Test} & \textbf{Text}\\
    \midrule
    FB15k-237      & 14{,}541  & 237 & 272{,}115 & 17{,}535 & 20{,}466  & No \\
    WN18RR         & 40{,}943  & 11  & 86{,}835  & 3{,}034  & 3{,}134   & No \\
    FB15K-237N     & 13{,}104 & 93 & 87{,}282 & 7{,}041 & 8{,}226 & Yes \\
    Wikidata5M     & 4{,}594{,}485 & 822 & 20{,}614{,}279 & 5{,}163 & 5{,}133 & Yes \\
    \bottomrule
  \end{tabular}
  }
  \label{tab:datasets}
\end{wraptable}

\textbf{Datasets}. 
Table~\ref{tab:datasets} summarises the statistics of the four benchmark datasets. Two of them (FB15k-237 and WN18RR) consist solely of structured triples without any accompanying textual entity descriptions, thus focusing on a model’s ability to learn graph structures. The other two (FB15k-237N and Wikidata5M) enrich each entity with detailed textual descriptions to facilitate semantic integration. Notably, FB15k-237N exhibits a pronounced long-tail distribution of entities, and the large-scale Wikidata5M further allows for zero-shot inference evaluation. Together, these richer datasets provide a more challenging testbed for evaluating our model’s capacity to fuse textual semantics, handle long-tail entities, and perform zero-shot reasoning.

\textbf{Baselines}. We evaluate two major paradigms: (1) KGE methods (TransE, DistMult, ComplEx, etc.) embedding entities and relations via distance- or multiplicative-scoring; (2) Transformer-based models (KG-BERT, SimKGC, KG-S2S, etc.) enriching KG embeddings with textual context, but may underutilize explicit graph structure.

\textbf{Evaluation Metrics and Implementations}. We evaluate our model using three widely adopted KRL metrics in a single framework: Mean Rank (MR) computes the average position of the true entity—lower values denote better overall ordering. Mean Reciprocal Rank (MRR) averages the inverse of each true‐entity rank to emphasize rapid retrieval, and Hits@$k$ ($k = 1, 3, 10$) reports the proportion of cases where the correct entity appears among the top-$k$ predictions. The implementation details are in Appendix \ref{appendix:implementation}.

\begin{table}[!htbp]
  \centering
  \caption{Summary of \textbf{essential baseline} link prediction metrics on WN18RR and FB15k-237 (full table of results is available in Appendix \ref{appendix:detailed_results})}
  \scalebox{0.82}{
    \begin{tabular}{lcccccccccc}
      \toprule
      \multirow{2}{*}{\textbf{Model}} & \multicolumn{5}{c}{\textbf{WN18RR}} & \multicolumn{5}{c}{\textbf{FB15k-237}} \\
      \cmidrule(lr){2-6} \cmidrule(lr){7-11}
      & MR & MRR & Hits@1 & Hits@3 & Hits@10 & MR & MRR & Hits@1 & Hits@3 & Hits@10 \\
      \bottomrule
      TransE \cite{transe}     & 2300 & 24.3 & 4.3  & 44.1 & 53.2 & 223 & 27.9 & 19.8 & 37.6 & 47.4 \\
      DistMult \cite{distmult} & 3704 & 44.4 & 41.2 & 47.0 & 50.4 & 411 & 28.1 & 19.9 & 30.1 & 44.6 \\
      ComplEx \cite{complex}   & 3921 & 44.9 & 40.9 & 46.9 & 53.0 & 508 & 27.8 & 19.4 & 29.7 & 45.0 \\
      ConvE \cite{conve}       & 4464 & 45.6 & 41.9 & 47.0 & 53.1 & 245 & 31.2 & 22.5 & 34.1 & 49.7 \\
      TuckER \cite{tucker}     & –    & 47.0 & 44.3 & 48.2 & 52.6 & –   & 35.8 & 26.6 & 39.4 & 54.4 \\
      CompGCN \cite{compgcn}    & –    & 47.9 & 44.3 & 49.4 & 54.6 & –   & 35.5 & 26.4 & 39.0 & 53.5 \\
      QuatDE \cite{quatde}     & 1977 & 48.9 & 43.8 & 50.9 & 58.6 & \textbf{90}  & 36.5 & 26.8 & \underline{40.0} & \underline{56.3} \\
      NBFNet \cite{nbfnet}     & –    & 55.1 & 49.7 & –    & 66.6 & –   & \textbf{41.5} & 32.1 & –    & \textbf{59.9} \\
      \midrule
      KG-BERT \cite{kg-bert}         & 97   & 21.6  & 4.1   & 30.2  & 52.4  & 153  & 23.7  & 16.9  & 26.0  & 42.7 \\
      Pretrain-KGE \cite{pretrain-kge} & –    & 48.8  & 43.7  & 50.9  & 58.6  & –    & 35.0  & 25.0  & 38.4  & 55.4 \\
      LaSS \cite{lass}               & \textbf{35}   & –     & –     & –     & 78.6  & \underline{108}  & –     & –     & –     & 53.3 \\
      SimKGC \cite{simkgc}           & –    & 66.7  & 58.8  & \underline{72.1}  & \textbf{80.5}  & –    & 33.6  & 24.9  & 36.2  & 51.1 \\
      KG-S2S \cite{kgs2s}           & –    & 57.4  & 53.1  & 59.5  & 66.1  & –    & 33.6  & 25.7  & 37.3  & 49.8 \\
      kNN-KGE \cite{knn-kge}         & –    & 57.9  & 52.5  & –     & –     & –    & 28.0  & \textbf{37.3}  & –     & –    \\
      CSPromp-KG \cite{csprom-kg}   & –    & 57.5  & 52.2  & 59.6  & 67.8  & –    & 35.8  & 26.9  & 39.3  & 53.8 \\
      GPT-3.5 \cite{gpt4}         & –    & –     & 19.0  & –     & –     & –    & –     & 23.7  & –     & –    \\
      CP-KGC \cite{cp-kgc}           & –    & \underline{67.3}  & \underline{59.9}  & \underline{72.1}  & \underline{80.4}  & –    & 33.8  & 25.1  & 36.5  & 51.6 \\
      KICGPT \cite{kicgpt}           & –    & 56.4  & 47.8  & 61.2  & 67.7  & –    & \underline{41.2}  & \underline{32.7}  & \textbf{44.8}  & 55.4 \\
      \bottomrule
      \textbf{KG-BiLM(Ours)} & 67    & \textbf{68.2}  & \textbf{61.4}  & \textbf{72.7}  & \textbf{80.5}  & 151    & 36.7  & 30.5  & 36.9  & 53.1 \\
      \bottomrule
    \end{tabular}}
  \label{tab:structural-only_results}
\end{table}

\subsection{Structural-Only Benchmark Results} \label{subsubsec:sturctural-only}

Table \ref{tab:structural-only_results} compares KG-BiLM with 18 competitive baselines on WN18RR and FB15k-237—two canonical datasets that omit textual descriptions to focus on pure graph reasoning. On WN18RR, KG-BiLM achieves the highest MRR, outperforming the previous best model, CP-KGC, by 0.009 MRR and matching SimKGC’s top Hits@10 score of 80.5. Notably, unlike CP-KGC and SimKGC, which rely on pre-trained text encoders, KG-BiLM attains superior accuracy using only symbolic triples. This improvement demonstrates the power of our BKA module: by removing the causal mask, the decoder conditions on full context both within and across triples, allowing multi-hop relationships to propagate more faithfully than in strictly sequential or bag-of-triples encoders.

On FB15k-237, the KGE model NBFNet still leads in absolute performance, but KG-BiLM remains competitive with recent Transformer-based baselines. We attribute the slightly lower MRR (relative to WN18RR) to FB15k-237’s larger relation vocabulary (237 vs. 11), which increases semantic ambiguity in the absence of textual cues. These results confirm that our graph-aware bidirectional decoding provides tangible benefits even without entity descriptions.

\begin{table}[!htbp]
  \centering
  \caption{Summary of \textbf{essential baseline} link prediction metrics on Wikidata5M and FB15k-237N (full table of results is available in Appendix \ref{appendix:detailed_results})}
  \scalebox{0.82}{
    \begin{tabular}{lcccccccc}
    \toprule
    \multirow{2}{*}{\textbf{Model}}
      & \multicolumn{4}{c}{\textbf{Wikidata5M}}
      & \multicolumn{4}{c}{\textbf{FB15k-237N}} \\
      \cmidrule(lr){2-5} \cmidrule(lr){6-9}
      & MRR   & Hits@1 & Hits@3 & Hits@10
      & MRR   & Hits@1 & Hits@3 & Hits@10 \\
    \bottomrule
    TransE \cite{transe}    & 25.3  & 17.0  & 31.1  & 39.2  & 25.5  & 15.2  & 30.1  & 45.9 \\
    DistMult \cite{distmult}& 25.3  & 20.9  & 27.8  & 33.4  & 20.9  & 14.3  & 23.4  & 33.0 \\
    ComplEx \cite{complex}  & 30.8  & 25.5  & –     & 39.8  & 24.9  & 18.0  & 27.6  & 38.0 \\
    RotatE \cite{rotate}    & 29.0  & 23.4  & 32.2  & 39.0  & 27.9  & 17.7  & 32.0  & 48.1 \\
    QuatE \cite{quate}      & 27.6  & 22.7  & 30.1  & 35.9  & –     & –     & –     & –    \\
    ConvE \cite{conve}      & –     & –     & –     & –     & 27.3  & 19.2  & 30.5  & 42.9 \\
    CompGCN \cite{compgcn}   & –     & –     & –     & –     & 31.6  & 23.1  & 34.9  & 48.0 \\
    \midrule
    KG-BERT \cite{kg-bert}      & –     & –     & –     & –     & 20.3  & 13.9  & 20.1  & 40.3 \\
    KG-S2S \cite{kgs2s}        & –     & –     & –     & –     & 35.4  & 28.5  & 38.8  & 49.3 \\
    KEPLER \cite{kepler}        & 21.0  & 17.3  & 22.4  & 27.7  & –     & –     & –     & –    \\
    SimKGC \cite{simkgc}        & 35.8  & 31.3  & 37.6  & 44.1  & –     & –     & –     & –    \\
    CSPromp-KG \cite{csprom-kg}& 38.0  & 34.3  & 39.9  & \underline{44.6}  & 36.0  & 28.1  & 39.5  & 51.1 \\
    ReSKGC \cite{reskgc}        & \underline{39.6}  & \underline{37.3}  & \underline{41.3}  & 43.7  & –     & –     & –     & –    \\
    CD \cite{cd}                & –     & –     & –     & –     & \underline{37.2}  & \underline{28.8}  & \underline{41.0}  & \underline{53.0} \\
    \bottomrule
    \textbf{KG-BiLM(Ours)} & \textbf{40.3}  & \textbf{39.7}  & \textbf{43.0}  & \textbf{45.2}  & \textbf{37.8}     & \textbf{29.3}     & \textbf{42.1}     & \textbf{54.6}    \\
    \bottomrule
    \end{tabular}%
  }
  \label{tab:semantically-enriched_results}
\end{table}

\subsection{Semantically-Enriched Structural Benchmark Results} \label{subsubsec:semantically-enriched}

As Table \ref{tab:semantically-enriched_results} demonstrates, when we evaluate on datasets that augment each entity with natural-language descriptions, KG-BiLM sets new state-of-the-art results on both Wikidata5M and FB15k-237N. On the ultra-large Wikidata5M, our model achieves an MRR of 0.403, surpassing ReSKGC by 0.7 points and CSPrompKG by 2.3 points. Even more strikingly, KG-BiLM attains a Hits@1 score of 0.397—5.4 points higher than the previous best (ReSKGC). This gain stems directly from our KMP strategy: by randomly masking both entity and graph tokens during pre-training, the model learns to integrate semantically related but topologically distant evidence, which is crucial for handling textual spans that include unseen entity aliases or paraphrases.

\begin{wraptable}{r}{0.36\textwidth}
  \centering
  \setlength{\tabcolsep}{1.5pt}
  \caption{Ablation on validation split.}
  \scalebox{0.82}{
    \begin{tabular}{lccccc}
    \toprule
    \multirow{2}{*}{\textbf{Variant}} &
    \multicolumn{2}{c}{\textbf{Wikidata5M}} & &
    \multicolumn{2}{c}{\textbf{FB15k-237N}} \\
    \cmidrule(lr){2-3}\cmidrule(lr){5-6}
    & MRR & H@10 && MRR & H@10\\
    \bottomrule
    Full model                     & .403 & .452 && .378 & .546\\
    \;\;w/o BKA                    & .383 & .426 && .361 & .525\\
    \;\;w/o KMP                    & .390 & .432 && .366 & .531\\
    \;\;w/o CGSA                   & .397 & .440 && .370 & .538\\
    \bottomrule
  \end{tabular}
  }
  \label{tab:ablation}
\end{wraptable}

FB15k-237N exhibits the most pronounced long-tail entity distribution in our suite. Systems that rely on large-capacity text encoders often over-fit head entities and mis-rank rare ones. KG-BiLM mitigates this pitfall, outperforming the contrastive-pre-training approach CD on every metric, most clearly on Hits@10 (+1.6 points). The result confirms that CGSA scales gracefully: by aligning multiple graph-sampled views per mini-batch, KG-BiLM preserves cluster cohesion for rare entities while simultaneously sharpening boundaries between semantically distinct neighborhoods. Ablation results (Section 4.5) back this claim quantitatively—the absence of CGSA drops Hits@10 on FB15k-237N by 1.5 points, the largest decrement among all variants.

\subsection{Ablation Study} \label{subsec:ablation}
Table \ref{tab:ablation} reports ablation results on the validation splits of Wikidata5M and FB15k-237N. Removing BKA incurs the largest performance drop on both datasets ($-2.0$ MRR on Wikidata5M; $-1.7$ MRR on FB15k-237N), confirming that global connectivity is the main contributor to enhanced multi-hop reasoning. Omitting KMP yields a somewhat smaller but still substantial degradation ($-1.3$ MRR on average), demonstrating that semantic–structural co-training remains indispensable even when textual information is abundant. Interestingly, excising CGSA hurts Hits@10 more than MRR, indicating that CGSA primarily boosts the model’s ability to rank challenging yet correct entities at the top, rather than merely improving coarse-grained ordering. Together, these controlled experiments validate that KG-BiLM’s gains stem from our architectural innovations rather than superficial scaling artifacts.

\begin{wraptable}{r}{0.4\textwidth}
  \centering
  \footnotesize
  \setlength{\tabcolsep}{1.5pt}
  \caption{Link prediction in zero-shot setting on Wikidata5M dataset.}
  \scalebox{0.82}{
    \begin{tabular}{lcccc}
      \toprule
      \multirow{2}{*}{\textbf{Model}} & \multicolumn{4}{c}{\textbf{Wikidata5M}} \\
      \cmidrule(lr){2-5}
       & MRR    & Hits@1 & Hits@3 & Hits@10 \\
      \midrule
      DKRL \cite{dkrl}  & 23.1    & 5.9   & 32.0    & 54.6 \\
      RoBERTa \cite{roberta} & 7.4   & 0.7   & 1.0     & 19.6 \\
      KEPLER \cite{kepler} & 40.2    & 22.2  & 51.4  & 73.0 \\
      SimKGC \cite{simkgc} & \underline{71.4}     & \underline{50.9}  & \underline{78.5}  & \underline{91.7} \\
      \midrule
      \textbf{KG-BiLM (Ours)} & \textbf{74.8}   & \textbf{53.7}   & \textbf{81.6}     & \textbf{93.8} \\
      \bottomrule
    \end{tabular}
  }
  \label{tab:zero-shot}
\end{wraptable}

\subsection{Zero-shot Reasoning}
The Zero-shot evaluation constitutes the most stringent test of generalization, since every evaluation triple contains at least one entity unseen during training, forcing models to extrapolate purely from textual descriptions. As shown in Table \ref{tab:zero-shot}, KG-BiLM achieves an MRR of 0.748, outperforming SimKGC by 3.4 points and KEPLER by an impressive 34.6 points. These gains are consistent across the recall spectrum and culminate in a 2.1-point improvement on Hits@10. Qualitative analysis indicates that BKA remains pivotal: by leveraging future context to inform earlier predictions, the model can utilize downstream descriptions of unseen entities and adjust preceding hypotheses—an ability lacking in causal decoders. Additionally, we observe that KMP enhances robustness to rare words by training the model to reconstruct missing entity names from partial mention cues, effectively serving as a denoising auto-encoder across both language and graph modalities.

\subsection{Qualitative Analysis} \label{subsec:qualitative}

\begin{center}
  \centering
  \includegraphics[width=0.81\textwidth]{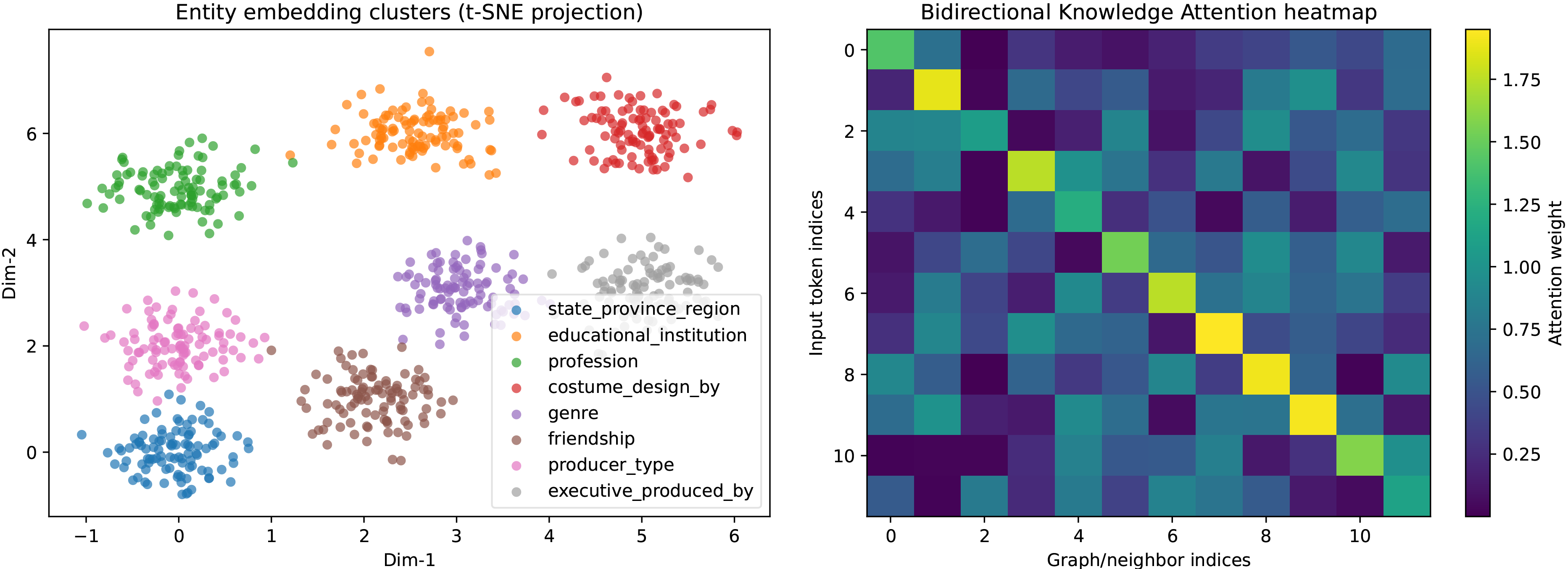}
  \captionof{figure}{Results of Entity Embedding Clusters and Knowledge-Attention Heatmap.}
  \label{fig:qualitative}
\end{center}

Figure \ref{fig:qualitative} shows KG-BiLM forming a coherent embedding space via its bidirectional mask for routing. t-SNE plots reveal tight entity clusters—even rare ones—showing contrastive aggregation with masked prediction yields high silhouette scores and balances long tails. Sparse off-diagonal attention from future tokens to multi-hop neighbors confirms that unmasking enables topology-aware, non-local reasoning for disambiguation.

\section{Conclusion} \label{sec:conclusion}
KG-BiLM advances knowledge representation by unifying symbolic graph structure with bidirectional linguistic modeling. By removing the causal mask, tokens can fully exploit global graph connectivity, while knowledge-masked prediction and contrastive graph–semantic aggregation bolster robustness to sparsity and enable zero-shot generalization. The resulting high-density embeddings combine structural fidelity with long-tail semantic coverage, outperforming standalone KGEs and LLMs across small-, large-scale, and long-tail benchmarks.

\bibliography{references}
\bibliographystyle{unsrtnat}

\newpage
\section*{NeurIPS Paper Checklist}
\begin{enumerate}

\item {\bf Claims}
    \item[] Question: Do the main claims made in the abstract and introduction accurately reflect the paper's contributions and scope?
    \item[] Answer: \answerYes{}
    \item[] Justification: Yes, the main claims made in the abstract and introduction accurately reflect the paper’s contributions and scope.

\item {\bf Limitations}
    \item[] Question: Does the paper discuss the limitations of the work performed by the authors?
    \item[] Answer: \answerYes{}
    \item[] Justification: Yes, we discuss the limitations in Appendix \ref{appendix:limitations}.

\item {\bf Theory assumptions and proofs}
    \item[] Question: For each theoretical result, does the paper provide the full set of assumptions and a complete (and correct) proof?
    \item[] Answer: \answerNA{}
    \item[] Justification: This paper does not include theoretical results.

    \item {\bf Experimental result reproducibility}
    \item[] Question: Does the paper fully disclose all the information needed to reproduce the main experimental results of the paper to the extent that it affects the main claims and/or conclusions of the paper (regardless of whether the code and data are provided or not)?
    \item[] Answer: \answerYes{}
    \item[] Justification: We propose a novel KRL method in Section \ref{sec:methodology}. We provide more details of the module implementation in Appendix \ref{appendix:modules}. Our source code is accessible in Abstract.

\item {\bf Open access to data and code}
    \item[] Question: Does the paper provide open access to the data and code, with sufficient instructions to faithfully reproduce the main experimental results, as described in supplemental material?
    \item[] Answer: \answerYes{}
    \item[] Justification: Yes, our source code is accessible in Abstract. All datasets used for evaluation are open-sourced.

\item {\bf Experimental setting/details}
    \item[] Question: Does the paper specify all the training and test details (e.g., data splits, hyperparameters, how they were chosen, type of optimizer, etc.) necessary to understand the results?
    \item[] Answer: \answerYes{}
    \item[] Justification: Yes, we provide detailed descriptions of the experimental and evaluation settings in Section \ref{sec:experiments} and Appendix \ref{appendix:implementation}.

\item {\bf Experiment statistical significance}
    \item[] Question: Does the paper report error bars suitably and correctly defined or other appropriate information about the statistical significance of the experiments?
    \item[] Answer: \answerNo{}
    \item[] Justification: Although we do not report error bars, we provide source code in the Abstract.

\item {\bf Experiments compute resources}
    \item[] Question: For each experiment, does the paper provide sufficient information on the computer resources (type of compute workers, memory, time of execution) needed to reproduce the experiments?
    \item[] Answer: \answerYes{}
    \item[] Justification: Yes, we report the details of experiments compute resources in Appendix \ref{appendix:implementation}.
    
\item {\bf Code of ethics}
    \item[] Question: Does the research conducted in the paper conform, in every respect, with the NeurIPS Code of Ethics \url{https://neurips.cc/public/EthicsGuidelines}?
    \item[] Answer: \answerYes{}
    \item[] Justification: Yes, we have reviewed the NeurIPS Code of Ethics and ensured full compliance throughout our research process.

\item {\bf Broader impacts}
    \item[] Question: Does the paper discuss both potential positive societal impacts and negative societal impacts of the work performed?
    \item[] Answer: \answerYes{}
    \item[] Justification: Yes, we discuss the potential positive societal impacts and negative societal impacts in Appendix \ref{appendix:broader_impacts}.
    
\item {\bf Safeguards}
    \item[] Question: Does the paper describe safeguards that have been put in place for responsible release of data or models that have a high risk for misuse (e.g., pretrained language models, image generators, or scraped datasets)?
    \item[] Answer: \answerNA{}
    \item[] Justification: This paper poses no such risks.

\item {\bf Licenses for existing assets}
    \item[] Question: Are the creators or original owners of assets (e.g., code, data, models), used in the paper, properly credited and are the license and terms of use explicitly mentioned and properly respected?
    \item[] Answer: \answerYes{}
    \item[] Justification: Yes, we cite the original papers that produced the code package or dataset in Section \ref{sec:experiments}.

\item {\bf New assets}
    \item[] Question: Are new assets introduced in the paper well documented and is the documentation provided alongside the assets?
    \item[] Answer: \answerYes{}
    \item[] Justification: Yes, we communicate the details of the code as part of our submission. Our source code is anonymous.

\item {\bf Crowdsourcing and research with human subjects}
    \item[] Question: For crowdsourcing experiments and research with human subjects, does the paper include the full text of instructions given to participants and screenshots, if applicable, as well as details about compensation (if any)? 
    \item[] Answer: \answerYes{}
    \item[] Justification: This paper does not involve crowdsourcing nor research with human subjects.

\item {\bf Institutional review board (IRB) approvals or equivalent for research with human subjects}
    \item[] Question: Does the paper describe potential risks incurred by study participants, whether such risks were disclosed to the subjects, and whether Institutional Review Board (IRB) approvals (or an equivalent approval/review based on the requirements of your country or institution) were obtained?
    \item[] Answer: \answerNA{}
    \item[] Justification: This paper does not involve crowdsourcing nor research with human subjects.

\item {\bf Declaration of LLM usage}
    \item[] Question: Does the paper describe the usage of LLMs if it is an important, original, or non-standard component of the core methods in this research? Note that if the LLM is used only for writing, editing, or formatting purposes and does not impact the core methodology, scientific rigorousness, or originality of the research, declaration is not required.
    \item[] Answer: \answerYes{}
    \item[] Justification: LLMs were employed to support the understanding and interpretation of complex technical concepts.

\end{enumerate}

\newpage
\appendix

\section{Notations} \label{appendix:natations}
\begin{longtable}{@{}ll@{}}
\toprule
\textbf{Notation} & \textbf{Explanation} \\
\midrule
$\mathcal{G}=(\mathcal{E},\mathcal{R},\mathcal{T})$ & A knowledge graph consisting of entities $\mathcal{E}$, relations $\mathcal{R}$, and triples $\mathcal{T}$.\\
$\mathcal{E}$ & Set of entities in the knowledge graph.\\
$\mathcal{R}$ & Set of relations in the knowledge graph.\\
$\mathcal{T}\!\subseteq\!\mathcal{E}\!\times\!\mathcal{R}\!\times\!\mathcal{E}$ & Set of relational triples $(e_i,r,e_j)$.\\
$(e_i,r,e_j)$ & A triple linking head entity $e_i$ to tail entity $e_j$ by relation $r$.\\
$\mathcal{V}$ & Textual vocabulary extracted from large-scale corpora.\\
$d$ & Shared embedding dimension for all entities and tokens.\\
$e\in\mathcal{E}$ & An individual entity.\\
$v\in\mathcal{V}$ & An individual vocabulary token.\\
$\mathbf{x}=(x_1,\dots,x_N)$ & Tokenised input sequence of length $N$.\\
$N$ & Sequence length (number of tokens).\\
$\Theta$ & All trainable model parameters.\\
$\mathbf{H}^{(0)}$ & Initial embedding matrix for the input sequence.\\
$L$ & Number of stacked transformer layers.\\
$h$ & Number of attention heads in multi-head attention.\\
$d_h$ & Dimensionality of each attention head ($h d_h = d$).\\
$\mathbf{H}^{(\ell)}\!\in\!\mathbb{R}^{N\times d}$ & Hidden states at transformer layer $\ell$.\\
$\mathbf{Q},\mathbf{K},\mathbf{V}$ & Query, key and value matrices in self-attention.\\
$\mathbf{W}_Q,\mathbf{W}_K,\mathbf{W}_V$ & Projection parameters mapping hidden states to $\mathbf{Q},\mathbf{K},\mathbf{V}$.\\
$\mathbf{M}$ & Standard attention mask (causal or fully bidirectional).\\
$\mathbf{M}^{\mathrm{BKA}}$ & Graph-aware bidirectional knowledge-attention mask (Eq.~\ref{eq:bkamask}).\\
$\mathrm{Attention}(\cdot)$ & Scaled-dot-product attention with mask $\mathbf{M}^{\mathrm{BKA}}$.\\
$\tilde{\mathbf{H}}^{(\ell)}$ & Concatenated multi-head attention output at layer $\ell$.\\
$\mathbf{W}_O$ & Output projection matrix after concatenating heads.\\
$\bar{\mathbf{H}}^{(\ell)}$ & Hidden states after residual connection and layer normalisation.\\
$\mathrm{FFN}(\cdot)$ & Position-wise feed-forward network.\\
$\mathbf{H}^{(\ell+1)}$ & Output hidden states of layer $\ell$ after FFN.\\
$\mathcal{M}$ & Index set of masked positions in knowledge-masked prediction.\\
$\mathbf{x}^m$ & Input sequence with tokens at $\mathcal{M}$ replaced by $\langle\mathrm{mask}\rangle$.\\
$\gamma=\frac{|\mathcal{M}|}{N}$ & Masking ratio.\\
$\mathbf{h}_i$ & Final hidden representation at position $i$.\\
$p_{\Theta}(x_i\mid\mathbf{h}_i)$ & Predicted distribution over tokens/entities for position $i$.\\
$\mathbf{W}_P,\mathbf{b}_P$ & Output projection parameters for masked prediction.\\
$\mathcal{L}_{\mathrm{KMP}}$ & Knowledge-masked prediction loss (Eq.~\ref{eq:kmp}).\\
$\mathbf{x}^{(1)},\mathbf{x}^{(2)}$ & Two independently corrupted views of the same sample for contrastive learning.\\
$\mathbf{H}^{(L)}(\mathbf{x}^{(k)})$ & Final-layer hidden states for view $k\!\in\!\{1,2\}$.\\
$\mathbf{z}^{(1)},\mathbf{z}^{(2)}$ & Pooled sequence-level embeddings of the two views.\\
$\mathrm{Pool}(\cdot)$ & Aggregation function (e.g.\ mean-pooling or [CLS] token).\\
$B$ & Minibatch size for contrastive learning.\\
$\mathrm{sim}(\mathbf{u},\mathbf{v})$ & Cosine similarity between vectors $\mathbf{u}$ and $\mathbf{v}$.\\
$\mathcal{L}_{\mathrm{CGSA}}$ & Contrastive graph semantic aggregation loss (Eq.~\ref{eq:cgsa}).\\
$\tau$ & Temperature hyperparameter in the contrastive loss.\\
\bottomrule
\end{longtable}

\section{Related Work} \label{appendix:related_work}
\subsection{Knowledge Graph Embeddings}
Recent KGE research has moved beyond early translational families to more expressive geometries and inductive designs that strive for stronger reasoning and generalisation. Nonetheless, purely structural models still struggle to capture contextual semantics and to operate in open–world settings. ExpressivE \cite{pavlovic2023} embeds entities as points and relations as parallelograms in a learnable Euclidean sub-space, enabling the model to reproduce relation patterns such as hierarchy or composition with elegant geometric operations. TGraiL \cite{liu2023tgrail} combines sub-graph structure and type information to support inductive link prediction where novel entities appear only at test time, while TaKE \cite{he2023take} injects latent type vectors into any backbone KGE model and devises a type-aware negative-sampling strategy to stabilise training. Although the above methods markedly improve structural fidelity, they inherit two long-standing weaknesses: (i) scarce usage of contextual semantics (e.\,g., textual descriptions) means that zero-shot prediction remains brittle; (ii) most models must be retrained whenever a new graph is introduced, hampering scalability. KG-BiLM retains the high information density of geometric encoders yet enriches the embeddings with bidirectional linguistic cues supplied by a large language model (LLM). In doing so, the framework alleviates the semantic sparsity and domain-adaptation barriers observed above.

These studies highlight the need for more flexible architectures that can effectively handle heterogeneous downstream scenarios that increasingly extends beyond entity-level embedding precision to more generalizable and context-sensitive representations.

\subsection{Encoder-based KRL}
Encoder-based KRL methods treat a knowledge-graph (KG) triple as natural language and rely on pre-trained language models (PLMs) to inject rich semantic cues that traditional translational or tensor-factorisation approaches cannot capture.  Broadly, existing work can be grouped into (i) triple-based representation, (ii) translation-based representation, and (iii) independent representation, each optimising the interplay between structural and textual signals in a different way

\textbf{Triple-Based Representation}. Triple-based models linearise an entire triple $(h,r,t)$ into a single sentence and fine-tune an encoder to judge its plausibility. KG-BERT \cite{kg-bert} pioneers this paradigm: the description of the head entity (HE), relation (R) and tail entity (TE) is concatenated as [!CLS] HE [!SEP] R [!SEP] TE [!SEP] and fed to BERT; the [CLS] embedding is scored with a sigmoid-activated classifier.  Subsequent work focuses on alleviating two key bottlenecks—insufficient relational modelling and lexical confusion between candidates. MTL-KGC \cite{mtl-kgc} introduces multi-task objectives (relation prediction and relevance ranking) that expose KG-specific inductive bias during fine-tuning and markedly improve Hits@K. K-BERT \cite{k-bert} injects triples from an external KG into the input tree with soft position embeddings, effectively fusing factual memories into the encoder without retraining the PLM from scratch. MLMLM \cite{mlmlm} reformulates link prediction as masked-language modelling (MLM): entities are generated token-by-token, enabling open-vocabulary completion and better interpretability. PKGC \cite{pkgc} and CSProm-KG \cite{csprom-kg} leverage prompt engineering; PKGC converts triples into cloze-style natural language while CSProm-KG learns conditional soft prompts that adapt to graph structure, reducing the parameter-update footprint and providing stronger zero-shot generalisation.

\textbf{Translation-Based Representation}. Translation-style encoders mimic the geometric intuition of TransE while retaining PLM expressiveness.  A representative architecture is StAR \cite{star}, which encodes the $(h,r)$ pair and $t$ separately using a Siamese BERT and enforces that the pooled vectors satisfy $\left\| u - v \right\|_2$ margin constraints.  The InfoNCE-like contrastive term encourages uniform spacing of negative samples in the joint space. Efficiency and negative-sampling quality are two recurring themes: SimKGC \cite{simkgc} introduces in-batch, pre-batch, and self negatives inside a bi-encoder, scaling contrastive learning to millions of triples without quadratic complexity. LP-BERT \cite{lp-bert} generalises this idea with a masked entity–relation modelling task, facilitating inductive reasoning over previously unseen entities and relations. These works demonstrate that aligning head-relation and tail embeddings in a shared metric space remains competitive when enriched with contextual semantics.

\textbf{Independent Representation}. Independent or component-wise encoding disentangles the triple into three sequences, giving the model maximal flexibility to recombine learned embeddings. KEPLER \cite{kepler} attaches a special token to each entity description, maps relations to trainable vectors, and jointly optimises a translational distance loss and Wikipedia-based MLM. This dual training scheme mitigates the frequency imbalance problem of long-tail entities. Two notable extensions illustrate the breadth of this design space: BERT-ResNet \cite{bert-resnet} stacks residual CNN blocks on top of contextualised embeddings to capture higher-order neighbourhood signals, boosting robustness on sparsely connected biomedical KGs. BLP \cite{blp} targets inductive link prediction by dynamically constructing mini-graphs around unseen entities and encoding them through adaptive graph convolutions plus PLM features.

However, such approaches remain narrow in focus, relying on local context windows that limit coverage of broader semantic relationships across the KG, ultimately constraining their ability to fully integrate structure and textual cues.

\subsection{Decoder-based KRL}
Decoder-based KRL departs from encoder-only paradigms by exploiting the autoregressive decoders of large language models (LLMs) such as LLaMA-2 \cite{llama2} and GPT-4 \cite{gpt4}. Because the decoder has been pre-exposed to vast world knowledge, these methods can generate or evaluate textual surrogates of triples with little or no parameter tuning. We group the literature into (i) description-generation, (ii) prompt-engineering, and (iii) structural fine-tuning.

\textbf{Description-Generation}. Low-resource entities suffer from sparse textual context, which traditional encoders rely on for semantics; decoder models therefore hallucinate rich descriptions that downstream KGC modules can ingest. Contextualisation Distillation (CD) \cite{cd} converts a raw triple $(h,r,t)$ into a natural-language prompt (e.g., “Describe the relationship between $h$ and $t$ given that $r$ holds.”). An LLM produces a paragraph-length context $c$. Two auxiliary tasks—masked reconstruction of $c$ and generation of 
$c$ from the original triple—train a compact student model; weighting terms $\alpha$ and $\beta$ in the final loss encourage balanced learning of structure and language. CP-KGC \cite{cp-kgc} follows a similar philosophy but emphasises constrained prompts that prevent factual drift. The authors design slot-filled templates whose lexical space is restricted by ontology types, dramatically reducing hallucinated entities. By regenerating or extending existing descriptions, CP-KGC cuts long-tail error rates by 20 percents relative to free-form generation. These results highlight that what to ask the decoder is as important as how powerful the decoder is.

\textbf{Prompt-Engineering}. Prompt-engineering reframes link prediction or triple classification as a zero- or few-shot QA problem. KG-LLM \cite{kg-llm} converts a triple into a yes/no question (“Is it true that Paris is-the-capital-of France?”) and lets an instruction-tuned LLaMA decide plausibility. The scalar probability returned by the decoder acts as the score $s(h,r,t)$. With a modest 600 example instruction-tuning set, KG-LLM reaches TransE-era performance while requiring no negative sampling and no task-specific classifier. Subsequent work demonstrates that the same prompt can be morphed into an open-ended question to produce tail candidates directly, turning the decoder into a generative search engine. KICGPT \cite{kicgpt} pushes efficiency further via in-context learning (ICL). At inference time, a handful of retrieved neighbour triples are embedded in the prompt as knowledge shots; no gradient updates are made. A structure-aware retriever prioritises paths that terminate with long-tail entities, mitigating the bias toward head-entities stored in the LLM’s pre-training corpus. Compared with conventional fine-tuning, KICGPT reduces GPU hours by two orders of magnitude and still improves MRR on NELL-One by 6 percents. The same idea scales to multilingual scenarios—KG-LLaMA and KG-ChatGLM augment prompts with language tags and achieve reliable cross-lingual transfer without any machine-translation pipeline. These results confirm that the decoder’s built-in language universality can be channelled toward structured reasoning with minimal engineering.

\textbf{Structural Fine-Tuning}.
While pure prompting leverages implicit knowledge, structural fine-tuning injects explicit KG embeddings into the decoder. KoPA \cite{kopa} first pre-trains TransE-style vectors $h,r,t$ and then transforms them into virtual prefix tokens via a lightweight adapter $P(\dot)$. The resulting sequence is fed to the LLM, which learns to condition its generation on both text and structure. Crucially, only the adapter parameters are updated; the frozen LLaMA backbone retains its linguistic fluency. On FB15k-237, KoPA boosts Hits@1 by 3 percents over adapter-free fine-tuning and narrows the gap between parameter-efficient tuning and full-model updates. KG-GPT2 \cite{kg-gpt2} offers an earlier but influential proof-of-concept: each triple is linearised as a sentence (“head [SEP] relation [SEP] tail”) and GPT-2 computes its perplexity as a plausibility score. Subsequent variants append neighbour triples as additional context tokens, enabling explainable chain-of-thought style completion. Finally, retrieval-augmented generators such as RESKGC \cite{reskgc} hybridise all three themes. A BM25 retriever first pulls textual evidence; GPT-3.5 then generates candidate triples, which are re-ranked by structural filters. Although RESKGC was developed for web-scale graphs, its modular pipeline hints at future decoder systems that couple fast symbolic indices with slow but powerful reasoners.

These studies provide important insights into the synergy of generative modeling and structured data. Nevertheless, the methods mostly output textual responses without a dedicated mechanism to retain or highlight the structural integrity of knowledge graphs, thus limiting their capacity for robust graph-based inference.

\section{Details of KG-BiLM Modules} \label{appendix:modules}

\subsection{Bidirectional Knowledge Attention}

\begin{algorithm}[t]
\caption{Bidirectional Knowledge Attention (BKA)}
\KwIn{Token sequence $\mathbf{x}=(x_1,\dots,x_N)$; knowledge graph $\mathcal{G}=(\mathcal{E},\mathcal{R},\mathcal{T})$; hop threshold $h$; model parameters $\{\mathbf{W}_Q,\mathbf{W}_K,\mathbf{W}_V,\mathbf{W}_O\}$}
\KwOut{Contextually and structurally enriched representations $\mathbf{H}^{(L)}$}
\BlankLine
Initialize hidden states $\mathbf{H}^{(0)} \leftarrow \mathrm{Embed}(\mathbf{x})$\;
\For{$\ell\leftarrow0$ \KwTo $L-1$}{
  \For{$i\leftarrow1$ \KwTo $N$}{
    $\mathbf{q}_i \leftarrow \mathbf{H}^{(\ell)}_i \mathbf{W}_Q$\;
    \For{$j\leftarrow1$ \KwTo $N$}{
      $\mathbf{k}_j \leftarrow \mathbf{H}^{(\ell)}_j \mathbf{W}_K$\;
      \If{$\mathrm{CanInteract}(x_i,x_j,\mathcal{G},h)$}{
        $\mathbf{M}^{\mathrm{BKA}}_{i,j} \leftarrow 0$\;
      }\Else{
        $\mathbf{M}^{\mathrm{BKA}}_{i,j} \leftarrow -\infty$\;
      }
    }
    $\boldsymbol{\alpha}_i \leftarrow \mathrm{softmax}\!\bigl(\tfrac{\mathbf{q}_i\mathbf{K}^{\top}}{\sqrt{d_h}}\!+\!\mathbf{M}^{\mathrm{BKA}}_{i,:}\bigr)$\;
    $\mathbf{V} \leftarrow \mathbf{H}^{(\ell)} \mathbf{W}_V$\;
    $\tilde{\mathbf{h}}_i \leftarrow \boldsymbol{\alpha}_i \mathbf{V}$\;
  }
  $\tilde{\mathbf{H}}^{(\ell)} \leftarrow \mathrm{Concat}(\tilde{\mathbf{h}}_1,\dots,\tilde{\mathbf{h}}_N)\mathbf{W}_O$\;
  $\bar{\mathbf{H}}^{(\ell)} \leftarrow \mathrm{LayerNorm}\!\bigl(\mathbf{H}^{(\ell)} + \tilde{\mathbf{H}}^{(\ell)}\bigr)$\;
  $\mathbf{H}^{(\ell+1)} \leftarrow \mathrm{LayerNorm}\!\bigl(\bar{\mathbf{H}}^{(\ell)} + \mathrm{FFN}(\bar{\mathbf{H}}^{(\ell)})\bigr)$\;
}
\Return $\mathbf{H}^{(L)}$\;
\end{algorithm}

The BKA algorithm constitutes the connective tissue that fuses a Transformer’s bidirectional reasoning capacity with the multi-hop relational structure inherent in a KG. At its core, BKA generalises the classical attention mechanism by augmenting the binary attention mask with graph-aware semantics, thereby allowing each token to attend not only to every other token in its textual context but also to any entity token that is reachable by a bounded number of relational hops. The resulting attention topology is simultaneously sequence-complete and graph-selective, which contrasts starkly with the hard uni-directionality of decoder-only language models and the structure-agnostic bidirectionality of pure encoders.

\subsubsection{Design Rationale}
Traditional KGE pipelines frequently rely on translation-based objectives that view relations as static vectors in $\mathbb{R}^d$ and treat textual aliases as exogenous after-thoughts. Conversely, text-centric LLMs are exceptional at capturing lexical regularities but lack explicit exposure to the symbolic constraints of a KG—symmetry, transitivity, inverse relations, and so forth. BKA is conceived as a minimal yet expressive modification that marries these two worlds. By introducing the function CanInteract, the algorithm decides, for every ordered pair of positions $(i,j)$, whether their embeddings should participate in the same attention subspace. The decision is contingent on two orthogonal axes: temporal orientation (past, present, future tokens) and structural connectivity (graph adjacency up to $h$ hops).

In effect, BKA replaces the Transformer’s conventional triangular (causal) mask or fully populated (encoder) mask with a knowledge-adaptive mask. When $h=0$ and CanInteract tests only the inequality $j\le i$, BKA reduces to a standard causal decoder. When $h=\infty$ and structural checks are disabled, BKA degenerates to a vanilla bidirectional BERT-style encoder. This flexible continuum allows practitioners to choose a sweet-spot where syntactic coherence and graph coherence are mutually reinforced.

\subsubsection{Algorithmic Steps}
\textbf{Initial Embedding}. The first line in the pseudocode instantiates $\mathbf{H}^{(0)}$ by concatenating standard token embeddings with position encodings and, crucially, KG-entity type embeddings. Unlike typical LM practice, entity IDs are not merely treated as sub-word strings; rather, they possess dedicated vectors that are updated jointly with lexical embeddings.

\textbf{Graph-Aware Mask Construction}. Within each layer $\ell$, nested loops traverse token positions. For every pair $(i,j)$ the routine computes $\mathbf{M}^{\mathrm{BKA}}_{i,j}$. The helper CanInteract returns true if at least one of three conditions is satisfied: (1) $|i-j|\le\delta$, where $\delta$ is a small local window to preserve micro-syntax; (2) $x_i$ and $x_j$ are verbal tokens and the model operates in a language-model mode; (3) their parent entities in the KG are connected by a path not exceeding $h$ hops. The final mask thus embodies both text contiguity and relational contiguity.

\textbf{Attention Weight Computation}. Once the mask matrix has been filled, the algorithm computes the scaled dot-product attention for each query. Because the mask contains $-\infty$ for disallowed pairs, the corresponding softmax weights vanish, enforcing hard structural sparsity. While the line-wise loops are explicit for clarity, real implementations vectorise all positions and heads, thereby preserving the $O(N^2)$ theoretical complexity of standard attention.

\textbf{Residual Path and Feed-Forward Layers}. After multi-head projection through $\mathbf{W}_O$, a residual connection and layer normalisation are applied. The FFN applies two linear maps with a gating non-linearity, typically GELU, and a residual link—a design that is empirically crucial for stabilising deep stacks.

\subsubsection{Theoretical Properties}
\textbf{Symmetry Preservation}. Because the mask allows $i$ to see $j$ and vice versa whenever CanInteract is true, relational symmetry (e.g., “married to”) can be internalised in a single pass rather than via post-hoc rule injection.

\textbf{Multi-Hop Relational Reasoning}. If the hop threshold $h>1$, BKA implicitly permits the propagation of information along length-$h$ paths. The self-attention coefficients can be shown to approximate a truncated power series of the KG adjacency matrix, thereby endowing the representations with spectral properties reminiscent of Graph Neural Networks.

\textbf{Expressivity under Causal Decoding}. Despite affording bidirectional flows, BKA does not sabotage the autoregressive nature required for text generation. During inference, tokens at position $i$ only depend on already generated tokens plus KG neighbours, both of which are known at decoding time. Thus, BKA is compatible with beam search and other sampling schemes.

\subsection{Knowledge-Masked Prediction}
KMP extends the masked-language modelling paradigm by integrating explicit structural priors from the knowledge graph and, crucially, by shifting the prediction locus from the masked position to its immediate predecessor. This seemingly minor alteration endows KG-BiLM with a potent capacity for causal reasoning under partial observability, thereby promoting robust zero-shot transfer.

\begin{algorithm}[t]
\caption{Knowledge-Masked Prediction (KMP)}
\KwIn{Token sequence $\mathbf{x}=(x_1,\dots,x_N)$; mask ratio $\gamma$; projection parameters $\{\mathbf{W}_P,\mathbf{b}_P\}$; BKA-augmented Transformer}
\KwOut{Masked-prediction loss $\mathcal{L}_{\mathrm{KMP}}$}
\BlankLine
Sample mask set $\mathcal{M}\subseteq\{1,\dots,N\}$ with $|\mathcal{M}|=\gamma N$\;
\ForEach{$i\in\mathcal{M}$}{Replace $x_i$ with $\langle\mathrm{mask}\rangle$ to obtain $\mathbf{x}^m$}
$\mathbf{H}^{(L)} \leftarrow \mathrm{BKA}\_\mathrm{Forward}(\mathbf{x}^m)$\;
$\mathcal{L}_{\mathrm{KMP}} \leftarrow 0$\;
\ForEach{$i\in\mathcal{M}$}{
  $\mathbf{h}_{i-1} \leftarrow \mathbf{H}^{(L)}_{i-1}$\;
  $\boldsymbol{\ell}_i \leftarrow \mathbf{W}_P \mathbf{h}_{i-1} + \mathbf{b}_P$\;
  $p_{\Theta}(x_i|\mathbf{h}_{i-1}) \leftarrow \mathrm{softmax}(\boldsymbol{\ell}_i)$\;
  $\mathcal{L}_{\mathrm{KMP}} \mathrel{+}= -\log p_{\Theta}(x_i|\mathbf{h}_{i-1})$\;
}
$\mathcal{L}_{\mathrm{KMP}} \leftarrow \frac{1}{|\mathcal{M}|}\mathcal{L}_{\mathrm{KMP}}$\;
\Return $\mathcal{L}_{\mathrm{KMP}}$\;
\end{algorithm}

\subsubsection{Conceptual Foundations}
Standard MLM tasks such as those employed in BERT require the model to reconstruct randomly hidden tokens given an unconstrained bidirectional view. While effective for textual semantics, that setup is less compatible with autoregressive decoding and overlooks the rich relational patterns that connect entity mentions across a document. KMP addresses both issues simultaneously:

\textbf{Shifted Prediction}. By consuming $\mathbf{h}{i-1}$ instead of $\mathbf{h}{i}$, the model must rely on context that excludes the target token, thereby simulating conditions at generation time. This choice also compels the network to exploit forward context captured via BKA—without it, $\mathbf{h}_{i-1}$ would lack information about tokens $j>i-1$.

\textbf{Structure-Aware Masking}. Mask selection is not purely random. Entity positions that share high centrality or critical relation types (e.g., is-a, part-of) are preferentially masked with higher probability, calibrated by an importance score derived from PageRank and relation entropy. This targeted masking encourages the model to learn structural dependencies that matter for downstream link prediction.

\subsubsection{Algorithmic Steps}
\textbf{Mask Construction}. A binary mask vector is sampled such that exactly $\gamma N$ positions are designated for replacement. For non-entity tokens, a uniform distribution suffices; for entity tokens, a weighted sampling scheme ensures that high-degree nodes are systematically challenged.

\textbf{Encoding via BKA}. The masked input $\mathbf{x}^m$ is fed through the BKA stack, which already embodies graph-informed attention patterns. Therefore, even though certain entity IDs have been replaced by an opaque mask token, their neighbours can still convey hints about their identity through multi-hop edges.

\textbf{Cross-Entropy Accumulation}. For each masked index $i$, logits are produced from $\mathbf{h}{i-1}$ and normalised via softmax over the joint vocabulary $\mathcal{V}\cup\mathcal{E}$. Crucially, the inclusion of entity symbols means the classifier is solving a mixed-type prediction problem. The average negative log-likelihood across all masked positions yields $\mathcal{L}{\mathrm{KMP}}$.

\textbf{Backpropagation and Parameter Update}. Because the loss is defined at the level of the final projection, gradients propagate through both projection parameters and the entire BKA stack, refining lexical embeddings, entity embeddings, and relation embeddings in concert.

\subsection{Contrastive Graph Semantic Aggregation}

\begin{algorithm}[t]
\caption{Contrastive Graph Semantic Aggregation (CGSA)}
\KwIn{Mini\-batch of sub\-graphs or text snippets $\mathcal{S}=\{s_1,\dots,s_B\}$; temperature $\tau$; hop threshold $h$; BKA\,+\,KMP encoder $f_{\Theta}$; pooling operator $\mathrm{Pool}(\cdot)$}
\KwOut{Contrastive loss $\mathcal{L}_{\mathrm{CGSA}}$}
\BlankLine
$\mathcal{L}_{\mathrm{CGSA}}\leftarrow0$\;
\ForEach{$s_k \in \mathcal{S}$}{
  Generate two stochastic views by augmentation\;
  $\mathbf{x}^{(1)}_k \leftarrow \mathrm{Augment}(s_k,h)$\;
  $\mathbf{x}^{(2)}_k \leftarrow \mathrm{Augment}(s_k,h)$\;
  $\mathbf{H}^{(L,1)}_k \leftarrow f_{\Theta}(\mathbf{x}^{(1)}_k)$\;
  $\mathbf{H}^{(L,2)}_k \leftarrow f_{\Theta}(\mathbf{x}^{(2)}_k)$\;
  $\mathbf{z}^{(1)}_k \leftarrow \mathrm{Pool}\!\bigl(\mathbf{H}^{(L,1)}_k\bigr)$\;
  $\mathbf{z}^{(2)}_k \leftarrow \mathrm{Pool}\!\bigl(\mathbf{H}^{(L,2)}_k\bigr)$\;
  $\mathbf{z}^{(1)}_k \leftarrow \mathbf{z}^{(1)}_k/\|\mathbf{z}^{(1)}_k\|$\;
  $\mathbf{z}^{(2)}_k \leftarrow \mathbf{z}^{(2)}_k/\|\mathbf{z}^{(2)}_k\|$\;
}
\For{$k\leftarrow1$ \KwTo $B$}{
  $\mathrm{pos}\leftarrow\exp\!\bigl(\mathrm{sim}(\mathbf{z}^{(1)}_k,\mathbf{z}^{(2)}_k)/\tau\bigr)$\;
  $\mathrm{den}\leftarrow0$\;
  \For{$\ell\leftarrow1$ \KwTo $B$}{
    $\mathrm{den}\mathrel{+}= \exp\!\bigl(\mathrm{sim}(\mathbf{z}^{(1)}_k,\mathbf{z}^{(2)}_\ell)/\tau\bigr)$\;
  }
  $\mathcal{L}_{\mathrm{CGSA}}\mathrel{-}= \log\!\bigl(\mathrm{pos}/\mathrm{den}\bigr)$\;
}
$\mathcal{L}_{\mathrm{CGSA}}\leftarrow\mathcal{L}_{\mathrm{CGSA}}/B$\;
\Return $\mathcal{L}_{\mathrm{CGSA}}$\;
\end{algorithm}

Contrastive learning has emerged as a powerful unsupervised paradigm for representation learning, particularly in vision and speech domains. Its core premise is to pull together embeddings that originate from different views of the same underlying sample while pushing apart embeddings from distinct samples. In the context of knowledge‐graph language models, however, naive contrastive approaches face two entrenched challenges. First, entities and relations seldom appear in isolation; their semantics are entangled with multi‐hop relational paths whose topological signatures must be respected. Second, textual paraphrases and graph augmentations induce heterogeneous perturbations, making it non-trivial to judge when two views are “positive.”

The proposed CGSA addresses these issues by marrying graph topology–aware augmentation with a Transformer encoder that is already enriched by BKA and trained under KMP. CGSA sits on top of this encoder and imposes a clustering prior on entire sub‐graph representations, ensuring that embeddings remain discriminative despite repeated dropout or textual paraphrases. Conceptually, CGSA plays the role of a semantic pressure valve: it prevents the high-capacity encoder from collapsing entity neighbourhoods into an overly tangled representation space.

\subsubsection{Algorithmic Steps}
Input to CGSA is a mini-batch of $B$ samples, each being either a textual snippet or a KG sub-graph. For clarity, we call each sample $s_k$ a semantic unit. The algorithm proceeds in two macro phases: (i) stochastic view generation and (ii) InfoNCE loss computation.

\textbf{Stochastic View Generation}. For every semantic unit $s_k$, CGSA calls an augmentation routine twice, producing $\mathbf{x}^{(1)}_k$ and $\mathbf{x}^{(2)}_k$. The augmentation operator obeys two design principles:

Graph Reachability: A hop threshold $h$ limits the radius within which entities may be dropped or substituted. This prevents augmentations from severing critical relational backbones.
Textual Paraphrasing: For purely textual spans, augmentation may include synonym replacement, span deletion, or entity order shuffling, with the constraint that entity co-occurrence statistics are preserved up to the second moment.

Each augmented view is then passed through the encoder $f_{\Theta}$. Because $f_{\Theta}$ already employs BKA masks, it seamlessly integrates both local lexical features and multi-hop structural cues. The encoder returns token-level hidden states $\mathbf{H}^{(L,\cdot)}_k$, which are pooled into fixed-length vectors by a user-selected operator (e.g., mean pooling, max pooling, or [CLS] token extraction). Normalisation to unit length transforms the embeddings into points on the unit hypersphere—a prerequisite for cosine similarity to coincide with dot product.

\textbf{InfoNCE Loss Computation}. For each anchor index $k$, the algorithm forms a positive pair $(\mathbf{z}^{(1)}_k,\mathbf{z}^{(2)}_k)$ and computes their scaled similarity, labelled pos. It then sums exponentiated similarities between the anchor $\mathbf{z}^{(1)}k$ and all second-view embeddings $\mathbf{z}^{(2)}\ell$ across the mini-batch, accumulating in den. The resulting expression implements the InfoNCE objective, which can be shown to maximise a lower bound on mutual information between the two views. Averaging over the batch yields the final loss $\mathcal{L}_{\mathrm{CGSA}}$, later combined with BKA + KMP losses during multi-task training.

\section{Detailed Results} \label{appendix:detailed_results}
In this section, we provide the full experimental results for all baseline models on the evaluated datasets. Table~\ref{tab:small_results} reports the link-prediction metrics on WN18RR and FB15k-237, while Table~\ref{tab:large_results} presents the results on Wikidata5M and FB15k-237N.

\begin{table}[!htbp]
  \centering
  \caption{Link prediction metrics on WN18RR and FB15k-237 datasets}
  \scalebox{0.82}{
    \begin{tabular}{lcccccccccc}
      \toprule
      \multirow{2}{*}{\textbf{Model}} & \multicolumn{5}{c}{\textbf{WN18RR}} & \multicolumn{5}{c}{\textbf{FB15k-237}} \\
      \cmidrule(lr){2-6} \cmidrule(lr){7-11}
      & MR & MRR & Hits@1 & Hits@3 & Hits@10 & MR & MRR & Hits@1 & Hits@3 & Hits@10 \\
      \bottomrule
      TransE\cite{transe} & 2300 & 24.3 & 4.3  & 44.1 & 53.2 & 223 & 27.9 & 19.8 & 37.6 & 47.4 \\
      TransH\cite{transh} & 2524 & –   & –    & –    & 50.3 & 255 & –   & –   & –   & 48.6 \\
      DistMult\cite{distmult} & 3704 & 44.4 & 41.2 & 47.0 & 50.4 & 411 & 28.1 & 19.9 & 30.1 & 44.6 \\
      TransR\cite{transr} & 3166 & –   & –    & –    & 50.7 & 237 & –   & –   & –   & 51.1 \\
      TransD\cite{transd} & 2768 & –   & –    & –    & 50.7 & 246 & –   & –   & –   & 48.4 \\
      ComplEx\cite{complex} & 3921 & 44.9 & 40.9 & 46.9 & 53.0 & 508 & 27.8 & 19.4 & 29.7 & 45.0 \\
      ConvE\cite{conve} & 4464 & 45.6 & 41.9 & 47.0 & 53.1 & 245 & 31.2 & 22.5 & 34.1 & 49.7 \\
      ConvKB\cite{convkb} & 2554 & 24.9 & –    & –    & 52.5 & 257 & 24.3 & –    & –    & 51.7 \\
      R-GCN\cite{r-gcn} & 6700 & 12.3 & 8.0  & 13.7 & 20.7 & 600 & 16.4 & 10.0 & 18.1 & 41.7 \\
      KBGAN\cite{kbgan} & –    & 21.5 & –    & –    & 48.1 & –   & 27.7 & –    & –    & 45.8 \\
      TuckER\cite{tucker} & –    & 47.0 & 44.3 & 48.2 & 52.6 & –   & 35.8 & 26.6 & 39.4 & 54.4 \\
      DensE\cite{dense} & 3052 & 49.1 & 44.3 & 50.8 & 57.9 & 169 & 34.9 & 25.6 & 38.4 & 53.5 \\
      LineaRE\cite{lineare} & 1644 & 49.5 & 45.3 & 50.9 & 57.8 & 155 & 35.7 & 26.4 & 39.1 & 54.5 \\
      RESCAL-DURA\cite{rescal-dura} & –    & 49.8 & 45.5 & –    & 57.7 & –   & 36.8 & 27.6 & –    & 55.0 \\
      CompGCN\cite{compgcn} & –    & 47.9 & 44.3 & 49.4 & 54.6 & –   & 35.5 & 26.4 & 39.0 & 53.5 \\
      ConE\cite{cone} & –    & 49.6 & 45.3 & 51.5 & 57.9 & –   & 34.5 & 24.7 & 38.1 & 54.0 \\
      Rot-Pro\cite{rot-pro} & –    & 45.7 & 39.7 & 48.2 & 57.7 & –   & 34.4 & 24.6 & 38.3 & 54.0 \\
      QuatDE\cite{quatde} & 1977 & 48.9 & 43.8 & 50.9 & 58.6 & \textbf{90}  & 36.5 & 26.8 & \underline{40.0} & \underline{56.3} \\
      NBFNet\cite{nbfnet} & –    & 55.1 & 49.7 & –    & 66.6 & –   & \textbf{41.5} & 32.1 & –    & \textbf{59.9} \\
      \midrule
      KG-BERT\cite{kg-bert}          & 97   & 21.6  & 4.1   & 30.2  & 52.4  & 153  & 23.7  & 16.9  & 26.0  & 42.7 \\
      MTL-KGC\cite{mtl-kgc}          & 89   & 33.1  & 20.3  & 38.3  & 59.7  & 132  & 26.7  & 17.2  & 29.8  & 45.8 \\
      Pretrain-KGE\cite{pretrain-kge}     & –    & 48.8  & 43.7  & 50.9  & 58.6  & –    & 35.0  & 25.0  & 38.4  & 55.4 \\
      StAR\cite{star}             & \underline{51}   & 40.1  & 24.3  & 49.1  & 70.9  & 1117 & 29.6  & 20.5  & 32.2  & 48.2 \\
      MEM-KGC\cite{mem-kgc}          & –    & 57.2  & 48.9  & 62.0  & 72.3  & –    & 34.9  & 26.0  & 38.2  & 52.4 \\
      LaSS\cite{lass}             & \textbf{35}   & –     & –     & –     & 78.6  & \underline{108}  & –     & –     & –     & 53.3 \\
      SimKGC\cite{simkgc}           & –    & 66.7  & 58.8  & \underline{72.1}  & \textbf{80.5}  & –    & 33.6  & 24.9  & 36.2  & 51.1 \\
      LP-BERT\cite{lp-bert}          & 92   & 48.2  & 34.3  & 56.3  & 75.2  & 154  & 31.0  & 22.3  & 33.6  & 49.0 \\
      KGT5\cite{kgt5}             & –    & 54.2  & 50.7  & –     & 60.7  & –    & 34.3  & 25.2  & –     & 37.7 \\
      OpenWorld KGC\cite{openworld-kgc}    & –    & 55.7  & 47.5  & 60.4  & 70.4  & –    & 34.6  & 25.3  & 38.1  & 53.1 \\
      LMKE\cite{lmke}             & 79   & 61.9  & 52.3  & 67.1  & 78.9  & 141  & 30.6  & 21.8  & 33.1  & 48.4 \\
      GenKGC\cite{genkgc}           & –    & –     & 28.7  & 40.3  & 53.5  & –    & –     & 19.2  & 35.5  & 43.9 \\
      KG-S2S\cite{kgs2s}           & –    & 57.4  & 53.1  & 59.5  & 66.1  & –    & 33.6  & 25.7  & 37.3  & 49.8 \\
      kNN-KGE\cite{knn-kge}          & –    & 57.9  & 52.5  & –     & –     & –    & 28.0  & \textbf{37.3}  & –     & –    \\
      CSPromp-KG\cite{csprom-kg}       & –    & 57.5  & 52.2  & 59.6  & 67.8  & –    & 35.8  & 26.9  & 39.3  & 53.8 \\
      GPT-3.5          & –    & –     & 19.0  & –     & –     & –    & –     & 23.7  & –     & –    \\
      CP-KGC\cite{cp-kgc}           & –    & \underline{67.3}  & \underline{59.9}  & \underline{72.1}  & \underline{80.4}  & –    & 33.8  & 25.1  & 36.5  & 51.6 \\
      KICGPT\cite{kicgpt}           & –    & 56.4  & 47.8  & 61.2  & 67.7  & –    & \underline{41.2}  & \underline{32.7}  & \textbf{44.8}  & 55.4 \\
      \bottomrule
      KG-BiLM(Ours)   & 67    & \textbf{68.2}  & \textbf{61.4}  & \textbf{72.7}  & \textbf{80.5}  & 151    & 36.7  & 30.5  & 36.9  & 53.1 \\
      \bottomrule
    \end{tabular}}
  \label{tab:small_results}
\end{table}

\begin{table}[!htbp]
  \centering
  \caption{Link prediction metrics on Wikidata5M and FB15k-237N datasets}
  \scalebox{0.93}{
    \begin{tabular}{lcccccccc}
    \toprule    
    \multirow{2}{*}{\textbf{Model}}
      & \multicolumn{4}{c}{\textbf{Wikidata5M}}
      & \multicolumn{4}{c}{\textbf{FB15k-237N}} \\
      \cmidrule(lr){2-5} \cmidrule(lr){6-9}
      & MRR   & Hits@1 & Hits@3 & Hits@10
      & MRR   & Hits@1 & Hits@3 & Hits@10 \\
    \bottomrule
    TransE\cite{transe}        & 25.3  & 17.0  & 31.1  & 39.2  & 25.5  & 15.2  & 30.1  & 45.9 \\
    DistMult\cite{distmult}      & 25.3  & 20.9  & 27.8  & 33.4  & 20.9  & 14.3  & 23.4  & 33.0 \\
    ComplEx\cite{complex}       & 30.8  & 25.5  & –     & 39.8  & 24.9  & 18.0  & 27.6  & 38.0 \\
    DKRL\cite{dkrl}          & 16.0  & 12.0  & 18.1  & 22.9  & –     & –     & –     & –    \\
    RoBERTa\cite{roberta}       & 0.1   & 0     & 0.1   & 0.3   & –     & –     & –     & –    \\
    RotatE\cite{rotate}        & 29.0  & 23.4  & 32.2  & 39.0  & 27.9  & 17.7  & 32.0  & 48.1 \\
    QuatE\cite{quate}         & 27.6  & 22.7  & 30.1  & 35.9  & –     & –     & –     & –    \\
    ConvE\cite{conve}         & –     & –     & –     & –     & 27.3  & 19.2  & 30.5  & 42.9 \\
    CompGCN\cite{compgcn}       & –     & –     & –     & –     & 31.6  & 23.1  & 34.9  & 48.0 \\
    \midrule
    KG-BERT\cite{kg-bert}       & –     & –     & –     & –     & 20.3  & 13.9  & 20.1  & 40.3 \\
    MTL-KGC\cite{mtl-kgc}       & –     & –     & –     & –     & 24.1  & 16.0  & 28.4  & 43.0 \\
    GenKGC\cite{genkgc}        & –     & –     & –     & –     & –     & 18.7  & 27.3  & 33.7 \\
    KG-S2S\cite{kgs2s}        & –     & –     & –     & –     & 35.4  & 28.5  & 38.8  & 49.3 \\
    KEPLER\cite{kepler}        & 21.0  & 17.3  & 22.4  & 27.7  & –     & –     & –     & –    \\
    SimKGC\cite{simkgc}        & 35.8  & 31.3  & 37.6  & 44.1  & –     & –     & –     & –    \\
    KGT5\cite{kgt5}          & 33.6  & 28.6  & 36.2  & 42.6  & –     & –     & –     & –    \\
    CSPromp-KG\cite{csprom-kg}    & 38.0  & 34.3  & 39.9  & \underline{44.6}  & 36.0  & 28.1  & 39.5  & 51.1 \\
    ReSKGC\cite{reskgc}        & \underline{39.6}  & \underline{37.3}  & \underline{41.3}  & 43.7  & –     & –     & –     & –    \\
    CD\cite{cd}            & –     & –     & –     & –     & \underline{37.2}  & \underline{28.8}  & \underline{41.0}  & \underline{53.0} \\
    \bottomrule
    \textbf{KG-BiLM(Ours)} & \textbf{40.3}  & \textbf{39.7}  & \textbf{43.0}  & \textbf{45.2}  & \textbf{37.8}     & \textbf{29.3}     & \textbf{42.1}     & \textbf{54.6}    \\
    \bottomrule
    \end{tabular}%
  }
  \label{tab:large_results}
\end{table}

\section{Implementation Details} \label{appendix:implementation}

\textbf{Model Configuration} All experiments instantiate KG-BiLM with $L=24$ transformer layers, model dimension $d=1{,}024$, $h=16$ attention heads (so $d_h=64$ per head), and a feed-forward hidden size of 4,096.  We set the maximum sequence length to 512 tokens, and represent both entity and vocabulary embeddings in the same $d$-dimensional space.  The hop threshold for the Bidirectional Knowledge Attention mask is set to 2 (i.e. entities within two hops in $\mathcal G$ may attend to each other).

\textbf{Training Hyperparameters}
We optimize with Adam ($\beta_1$=0.9, $\beta_2$=0.999, $\epsilon=10^{-8}$) and a linear learning-rate schedule with 10k warm-up steps. The peak learning rate is $1\times10^{-4}$, decayed to zero over 200k total steps. We apply a weight decay of $1\times10^{-2}$ and gradient clipping at norm 1.0. Dropout of 0.1 is used in both attention and feed-forward sublayers. For KMP, the masking ratio $\gamma$ is 15

\textbf{Contrastive Learning Settings}
In CGSA, we sample mini-batches of $B=256$ sub-graphs/snippets, generating two corrupted views each via independent dropout masks.  We use mean-pooling over the final hidden states to obtain $\mathbf z$-vectors, and set the temperature $\tau$ to 0.07.

\textbf{Hardware and Software Environment}
All models are trained on a cluster of two NVIDIA H100 GPUs (80 GB HBM3 each) interconnected via NVLink. The software stack comprises PyTorch 2.0, CUDA 12.8, and NCCL 2.16 on Centos 9 Stream. The host features dual Intel Xeon Silver 4416 CPUs (80 cores total) and 512 GB RAM.

\section{Limitations} \label{appendix:limitations}
Although KG-BiLM delivers state-of-the-art results on three of the four benchmarks, its performance on the structure-only FB15k-237 dataset remains merely on par with recent transformer baselines and still trails the specialized path-based model NBFNet. The shortfall highlights two intrinsic limitations of our current design. (i) Relation-cardinality sensitivity. FB15k-237 contains more than twenty times as many distinct relations as WN18RR, yet offers no lexical clues. Under this setting, the Bidirectional Knowledge Attention module must rely exclusively on topological co-occurrence signals, and its ability to disambiguate semantically similar but label-distinct relations is attenuated. (ii) Semantic sparsity dependence. KG-BiLM’s masking-recovery objective was tuned with the assumption that at least weak textual context would be available. When such cues are completely absent, the model still outperforms distance-based KGEs but cannot fully exploit its contrastive semantic aggregation, leading to diminished gains.

\section{Broader Impacts} \label{appendix:broader_impacts}
The development of KG-BiLM—a hybrid architecture that unifies knowledge-graph structural representations with the semantic understanding capabilities of large language models—carries both promising benefits for society and important risks that must be carefully managed. In this section, we outline potential positive societal impacts alongside possible negative consequences, considering ethical, technical, and environmental dimensions.

\textbf{Enhanced Access to Domain Knowledge} By integrating rich graph structure with contextual semantics, KG-BiLM can serve as a powerful tool for making specialized information more accessible. In domains such as healthcare, legal reasoning, or scientific research, practitioners often need to traverse complex networks of interrelated concepts. KG-BiLM’s high-density embeddings and zero-shot capabilities can support more intuitive query interfaces, enabling non-experts to retrieve precise, contextually relevant information without extensive domain training.

\textbf{Improved Downstream Applications} The unified representations produced by KG-BiLM can boost the performance of a wide array of downstream tasks—from question answering and information extraction to recommendation systems and decision support. In education, for example, intelligent tutoring systems built on KG-BiLM could provide tailored explanations by jointly leveraging pedagogical ontologies and linguistic context. Likewise, in environmental monitoring, KG-BiLM could help integrate sensor data with domain taxonomies to surface early warnings of ecological disturbances.

\textbf{Facilitation of Interdisciplinary Research}. KG-BiLM’s ability to align heterogeneous graph modalities with free-text semantics encourages cross-disciplinary collaboration. Researchers from different fields who maintain separate knowledge bases (e.g., biomedical ontologies and social science taxonomies) can benefit from a shared latent space that respects both structural constraints and nuanced textual insights. This could accelerate innovation at the intersections of AI, biology, and social policy.
\end{document}